# Computer vision-based recognition of liquid surfaces and phase boundaries in transparent vessels, with emphasis on chemistry applications

*Sagi Eppel[a]\* and Tal Kachman[b]*

[a]Department of Materials Science and Engineering, Technion – Israel Institute of Technology, Haifa 32000, Israel.

[b]Physics Department, Technion-Israel Institute of Technology, Haifa 32000, Israel

E-mail: sagieppel@gmail.com
Tel: +972 523 202 516

## Abstract

The ability to recognize the liquid surface and the liquid level in transparent containers is perhaps the most commonly used evaluation method when dealing with fluids. Such recognition is essential in determining the liquid volume, fill level, phase boundaries and phase separation in various fluid systems. The recognition of liquid surfaces is particularly important in solution chemistry, where it is essential to many laboratory techniques (e.g., extraction, distillation, titration). A general method for the recognition of interfaces between liquid and air or between phase-separating liquids could have a wide range of applications and contribute to the understanding of the visual properties of such interfaces. This work examines a computer vision method for the recognition of liquid surfaces and liquid levels in various transparent containers. The method can be applied to recognition of both liquid-air and liquid-liquid surfaces. No prior knowledge of the number of phases is required. The method receives the image of the liquid container and the boundaries of the container in the image and scans all possible curves that could correspond to the outlines of liquid surfaces in the image. The method then compares each curve to the image to rate its correspondence with the outline of the real liquid surface by examining various image properties in the area surrounding each point of the curve. The image properties that were found to give the best indication of the liquid surface are the relative intensity change, the edge density change and the gradient direction relative to the curve normal.

# 1. Introduction

The visual identification of the liquid level and liquid surface in transparent vessels is perhaps the most commonly used analytical method for dealing with fluids. Such recognition has applications in a wide variety of fields, ranging from industry bottle-filling to everyday water and beverage handling (Figure 1).[1-32] One of the fields in which liquid surface recognition is most commonly used is solution chemistry. The ability to recognize liquid surfaces in the chemistry laboratory is essential for the estimation of liquid volume and liquid level as well as for the identification of phase boundaries and phase separation. The identification of such properties is essential to many laboratory techniques,[1,33] including liquid-liquid extraction,[3,31] column chromatography, distillation, titration and nearly any other method used in solution chemistry (Figure 1).

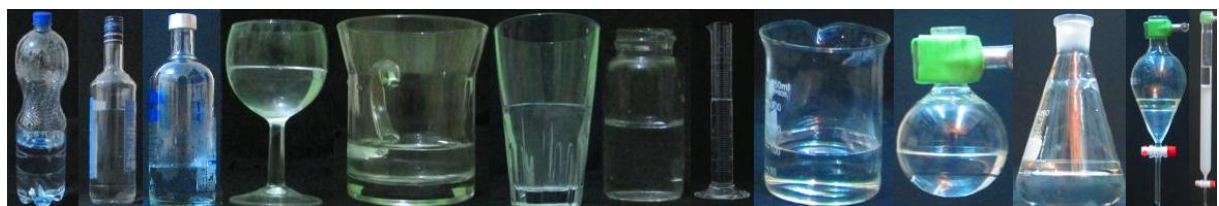

**Figure 1. Various pieces of axisymmetric liquid-containing glassware vessels used in a chemistry laboratory and for everyday purposes. From left to right: Plastic and glass bottles, kitchen cups, vial, graduated cylinder, beaker, boiling flask/round bottom flask, Erlenmeyer flask, separatory funnel, chromatography column.[33]**

To date, little research has been performed on using computer vision to recognize surfaces between liquid and air,[16,18-22] and even fewer works have focused on the recognition of surfaces between phase-separating liquids.[1,3,31] The methods available are either based on specific additives such as colored floating beads,[1,3,31] or assume specific conditions. A general system that can identify phase boundaries and surfaces with no additives and in uncertain conditions, similarly to a human observer, is not yet available. Such a system could potentially be used for the automation of many laboratory processes.[1,3,34-48] A general system that can determine the location of a liquid surface in transparent glassware could also be of use in any field in which liquid handling in a transparent vessel is necessary. In addition, the determination of the visual properties that characterize liquid surfaces can allow deeper understanding of the most commonly used and least explored analytical method in chemistry, that of vision.[1] In this paper, we suggest and demonstrate such a method that, while focused on chemical solvents and glassware, can be applied to any liquid in a transparent vessel. The method receives the image of the liquid and the boundaries of the liquid container in the image (Figure 2.b), then determines the location and shape of all liquid surfaces in the image.

No prior knowledge of the number of liquid phases or their properties is required. The only restriction regarding the vessel shape is that the vessel must be axisymmetric (cylindrical symmetry), which is by far the most common symmetry for a liquid container in any field. However, the method can be extended to other container symmetries. The method scans the vessel area in the image, line by line, and generates curves corresponding to the outlines of all possible liquid surfaces in the vessel (Section 2). The curves generated are then examined and rated according to their correspondence to the outlines of real liquid surfaces in the image. The curve rating is achieved by evaluating a specific property of the image around each point of the curve and using the result to rank the curve's correspondence with the liquid surface (Section 3). Various image properties such as the intensity change and the edge density have been evaluated as indicators for the liquid surface outline (Section 4). The best indicators were found to be the relative intensity change normal to the curve as well as the change in edge density and the gradient direction (Section 4). The main reasons for failed recognition and false recognitions are also examined (Sections 5-8).

## 1.1. Computer vision in the chemistry laboratory, hierarchical approach

Vision-based recognition is the most common type of analysis used in dealing with a chemical process.[1, 33] Despite the extensive use of vision-based recognition in the laboratory, only minor attention has been given to the use of computer vision in chemistry.[1, 3, 34, 39, 44, 47, 49-67] A contemporary review addressing the use of cameras and computer vision in chemistry by Ley and coworkers[1] suggests that little research exists thus far in this field. One of the few examples of computer vision in chemistry is the use of colored floating beads for tracking interfaces in liquid-liquid extraction.[3, 31, 35] The fact that this method has already been successfully applied in several automatic chemistry systems demonstrates the potential of machine vision in chemistry. Ideally, a computer vision system for chemistry will be able to replicate the ability of a human observer to identify the myriad aspects of chemical systems without assistance or additives. The main challenge in developing such a general computer vision system is the diversity of the problem. A complete visual analysis of any chemical system will first require the recognition of the vessel in which the chemical is stored, followed by the recognition of the phases within the vessel and finally the analysis of the properties of the phases. Solving all of these problems simultaneously requires a considerable amount of work and will result in a complex but not necessarily usable solution.

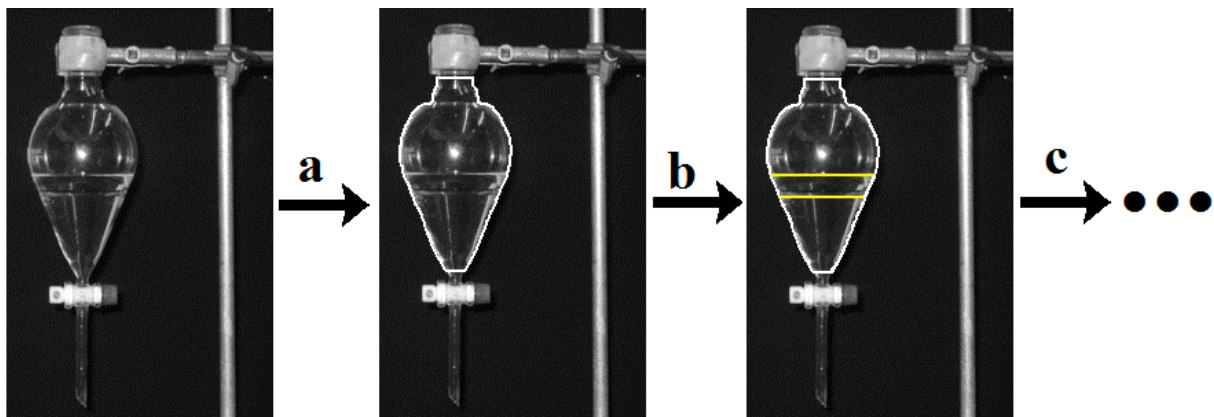

**Figure 2. Hierarchical layer approach to computer vision in the chemistry laboratory. Each layer performs one level of recognition, and then transfers the result to the next layer, which performs another level of recognition. a) First layer: recognition of the vessel borders. b) Second layer: recognition of the liquid and air phases within the vessel. c) Recognition and analysis of the properties and type of each phase (e.g., color, volume, emulsion).**

A simple way to avoid this problem is by using a hierarchical layer-by-layer approach. In this approach, each layer of recognition is performed independently, and the results are passed to the next layer (Figure 2). Each layer focuses on the recognition of one specific set of features, assuming that all necessary information has already been found by higher layers. The first layer recognizes the liquid vessel and traces its boundaries in the image (Figure 2.a). The first layer then passes the results to the second layer, which recognizes the liquid and gas phases in the vessel (Figure 2.b). The boundaries of these phases are transferred to the next layers, which can recognize further features (e.g., emulsion, solids, color). The advantage of the hierarchical approach is that it allows each step of the recognition to be addressed independently, which considerably simplifies the problem. The current work will address the second layer of recognition, the identification of the liquid phases. The boundaries of the vessel will be assumed to have been found in the previous layer. The recognition of the vessel boundaries can be achieved either by separating the vessel from the background or by using the template in the vessel shape (Figure 2.a). Source codes and instructions for performing vessel-boundaries recognition are available freely (See Supporting Information, section 9).

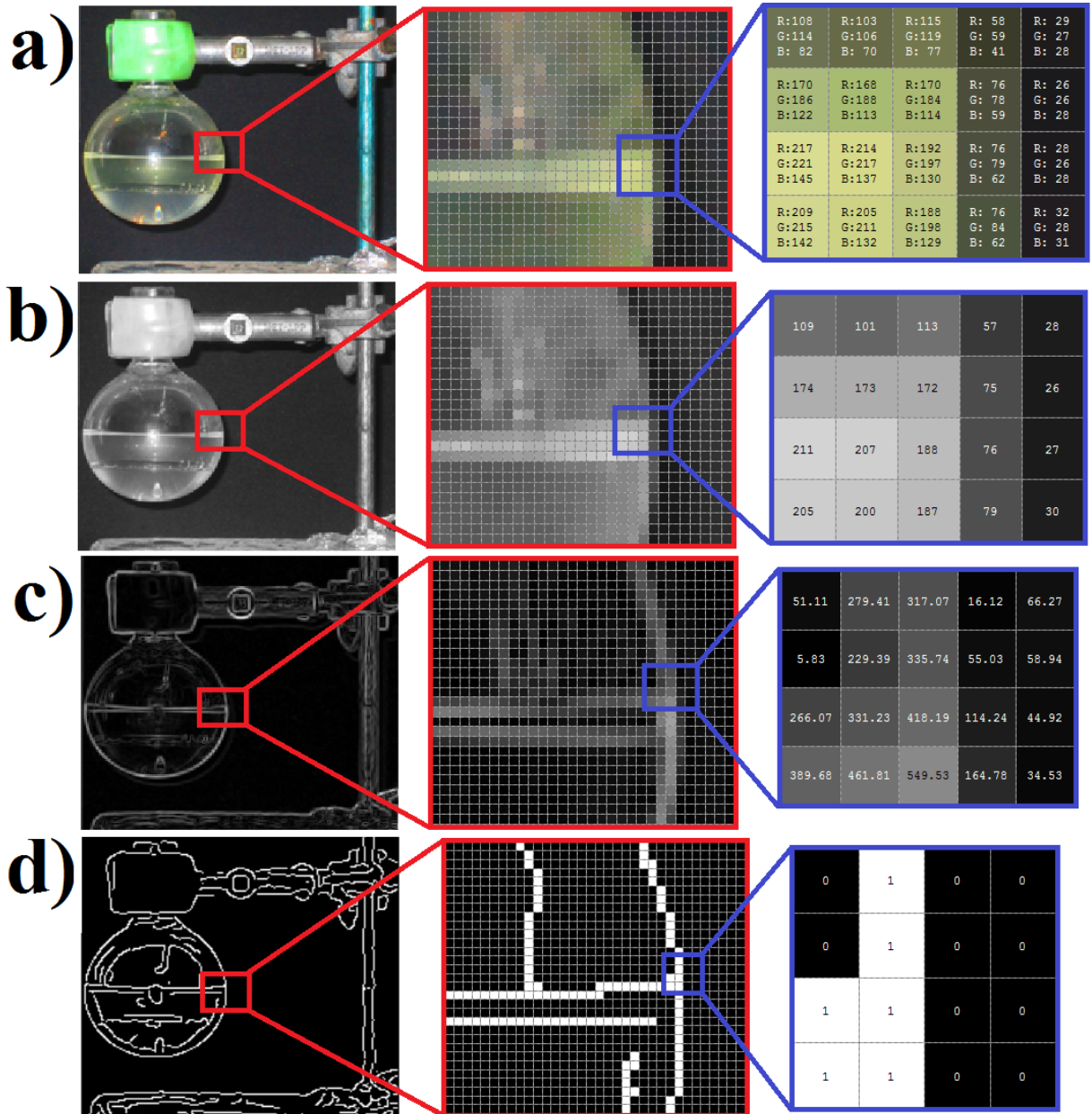

**Figure 3.** a) Color image represented as a 2D matrix. Each matrix cell (pixel) contains 3 parameters (R, G, B) that give the color saturation of Red, Green and Blue at the pixel location. b) The grayscale image is represented as a 2D matrix. Each matrix cell (pixel) has a single value that gives the gray level intensity in the pixel location. Intensity ranges from 0 (black) to 255 (white). c) Gradient size image: the value of each pixel is the size of the intensity gradient in this location. The gradient size is represented as the brightness level in the figure. d) Edge image: a binary image in which pixels corresponding to edges have a value of 1 (white), while all other pixels have a value of 0 (black).

## 1.2. Basic concepts in image analysis

This section will provide a brief review of some essential concepts in image analysis that will be used in this work.

**Pixel:** All digital images are written as a 2D matrix, where each cell in the matrix is referred to as a pixel. The value of each pixel represents the color of the pixel location in the image (Figure 3).[68, 69]

**Color image (RGB):** Each pixel in the RGB color image has three components (R, G, B).[68, 69] These components correspond to the saturation of the Red, Green and Blue colors in that pixel (Figure 3.a). Each component can have a value ranging from 0 to 255, where 255 means full color saturation (Figure 3.a). The combination of only three colors (wavelengths) with different intensities can generate the majority of the colors perceived by human vision. The RGB scheme is therefore used in the majority of digital screens, as well as for the storing and processing of color images.

**Grayscale image:** Grayscale images are essentially black-and-white images in which each pixel has a single value that represents the gray level or intensity in this pixel location (Figure 3.b).[68, 69] The value of each pixel can range from 0 (black) to 255 (white). For human vision, the colors of such images are perceived as shades of gray (Figure 3.b). Grayscale images are commonly used in computer vision because they can be analyzed as 2D matrices or 2D functions.

**Gradient image:** Gradient images are essentially the gradient maps of the intensity in the grayscale image (Figure 3.c).[68, 69] The value of each pixel in the gradient matrix represents the value of the intensity gradient in this pixel location. The gradient size is calculated as the Pythagorean sum of the difference between the intensity of the pixel and its close neighbors (in the grayscale image). The importance of the gradient map is that the edges of objects and patterns in images are usually characterized by a strong intensity change. Therefore, edges in the image are usually found by locating image areas with a large gradient (Figure 3.c). [68-71]

**Edge image:** The recognition of features in images is usually achieved by first finding the edges of the features.[68-71] Edge recognition is therefore an extremely important aspect of computer vision. An edge image is a binary matrix (Figure 3.d) in which each pixel (cell) can have one of two values: 0 (black) or 1 (white) (Figure 3.d). Pixels with values of 1 correspond to edges in the image (Figure 3.d). Edges in the image are usually characterized by a sharp change in intensity. Therefore, a simple method for the identification of edges is using all pixels in which the intensity gradient size exceeds some threshold value (Figure 3.c). While more sophisticated methods such as the canny operator[70] are usually used, the methods all use the intensity gradient magnitude as the main indicator for edge recognition.

### 1.3. Review of research in liquid level recognition

The available approaches on the machine vision-based recognition of liquid surfaces and phase boundaries could be divided into three major categories.

a) Recognition of the liquid level based on the combination of computer vision with external additives.[1-14] An example of the use of additives is the detection of liquid-liquid interfaces using a colored floating bead with a density between those of the upper and lower phases, which causes the bead to float directly on the interface.[1, 3, 4] Additives such as laser beams,[8-14] shaped light and shaped background patterns[5-7] have also been used. The additives are easily identified in the image. Their position and shape in the image are used to locate the liquid level.

b) Color-based recognition of liquids can be used in cases where the liquids have a distinct color. The recognition in this case is achieved by finding the region in the image that corresponds to the liquid color.[1, 19] By setting the background to a unique color, this method can in theory be applied to any non-transparent liquid.

c) Edge-based detection of the liquid level is perhaps the most efficient nonadditive computer vision approach when dealing with transparent fluids. This approach uses the edge image or intensity gradient (section 1.2) to identify the liquid line in the image.[15-18, 20-23] Few variations of this approach have been suggested. One approach is to identify the first line in which the vertical intensity gradient passes some threshold as the liquid level (Figure 3.c).[15] The second edge-based approach is to identify the longest horizontal edge line in the canny edge image as the liquid level (Figure 3.d).[22] Yet another edge-based approach is to take the average vertical distance from some reference line to the closest lower edge point in the edge image.[16, 18, 20]

### 1.4. General computer vision approach for the recognition of the liquid surfaces in a vessel

All the existing approaches for recognition of the liquid level have reported good results for the task for which they were designed. However, these methods are limited by either the use of specific additives or the need for a specific set of conditions.

A machine vision approach that will replicate the ability of a human observer to find the boundaries of liquid phases in an image must address four major challenges:

a) The various shapes of vessels in which the liquid can be stored. The large number of glassware shapes used in chemistry (Figure 1) makes this requirement particularly important for this field.

b) The liquid surface, which can take various shapes in the image. Current methods usually assume that the liquid phase boundary will take the shape of straight line. This assumption is true when the surface is viewed from a small angle (Figure 5.a). However, depending on the angle of view, the liquid surface shape can take various forms in the image (Figure 5).

c) Recognizing liquids with unknown properties such as color opacity, density and emulsion. This requirement is particularly important for the chemistry laboratory application, where large numbers of different materials are used, often with unknown properties.

d) Recognition of phase separation for systems with an unknown number of phases. The ability to identify phase separation as well as the phase boundaries is essential for many chemistry applications, including liquid-liquid extraction and column chromatography.

This work will suggest a general approach that confronts these challenges.

## 2. Scanning possible liquid surface curves

The liquid surface in an image is identified by scanning all curves that correspond to a possible outline of the liquid surface in a given vessel shape. The curves that overlap with the outline of the real liquid surface in the image are then identified. This approach is performed in three steps: a) Generate all curves that correspond to the possible outline of the liquid surface in a given vessel image (Section 2.1). b) For each curve, find the score that represents the curve match to the outline of the real liquid surface in the image (Section 2.2). c) Accept the curves with the best scores or the curves with the scores that pass some threshold (Section 2.3). A detailed description of each step is given in the following sections.

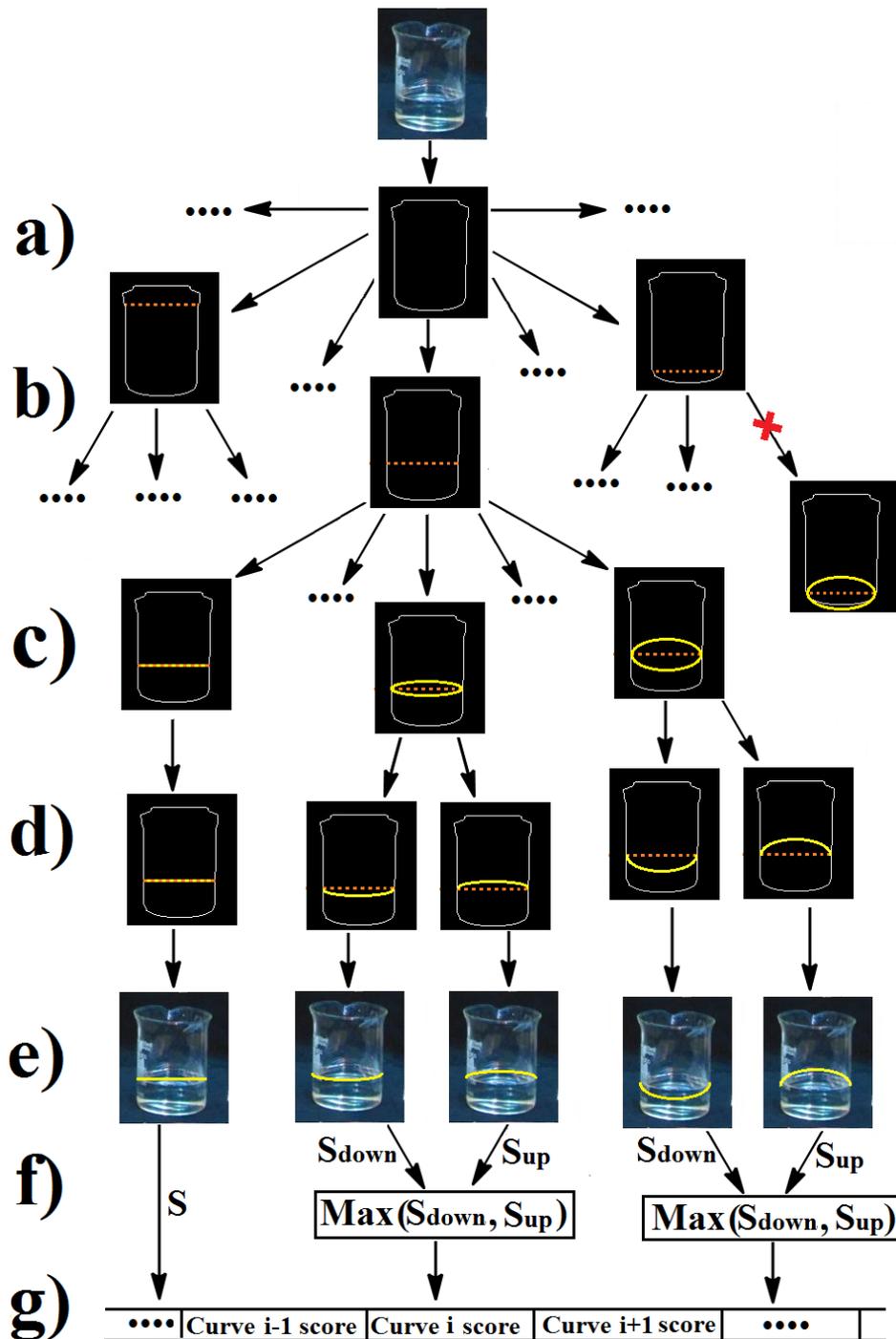

**Figure 4.** Scheme of the method for tracing the liquid surfaces in an image. a-b) Scan over all horizontal lines inside the vessel boundaries (in the image). c) For each line, generate all of the horizontal ellipses for which the current line is the major horizontal axis. The height of the ellipse can range from one pixel to some fraction of the ellipse width. If the curve thus generated exceeds the vessel boundaries, ignore it. d) Divide the ellipse into upper and lower curves. The upper curve contains all of the points of the ellipse above and on the center line. The lower curve contains all of the points on the ellipse on and below the center line (if the curve is a straight line, use the curve as it is). e) The upper and lower curves are each separately matched to the image. Each curve is given scores corresponding to the match between the curves and the liquid surface in the image. f) The higher score of the two curves is taken as the full ellipse score. g) The scores for all of the curves in all lines are registered. Curves with scores that exceed a minimal threshold are accepted as surface lines. Curves that are too close to curves with higher scores are removed.[72]

## 2.1. Generating all curves corresponding to possible liquid surfaces outlines

Generating all of the possible curves that could correspond to the outline of the liquid surface in a given image can result in very large numbers of possibilities, making such a scan impractical for most purposes. However, the number of possible curves can be dramatically reduced by assuming that the camera is not tilted to the left or the right. Under this assumption, all of the possible liquid surface curves could be generated by scanning all horizontal lines inside the vessel area in the image (Figure 4.b). For each line, all of the possible shapes of the liquid surfaces (centered on this line) are generated. For axisymmetric (cylindrical) vessels where the camera distance is larger than vessel radios, the liquid surface in the image will take the approximate shape of a straight line or a horizontal ellipse (Figure 4.c). The observed height of the ellipse depends on the angle of view (Figure 5.b). The liquid surface will appear as a straight line if viewed from very far or if the line of sight is the exact height of the surface (Figure 5.a). The liquid surface will appear as a full circle if the vessel is viewed directly from above (Figure 5.c). The identification of the liquid level in a transparent container is mostly performed from the side (Figure 5.b). As result, the maximum height of the ellipse that is generated could be restricted to some fraction of the line width (30% in this work).

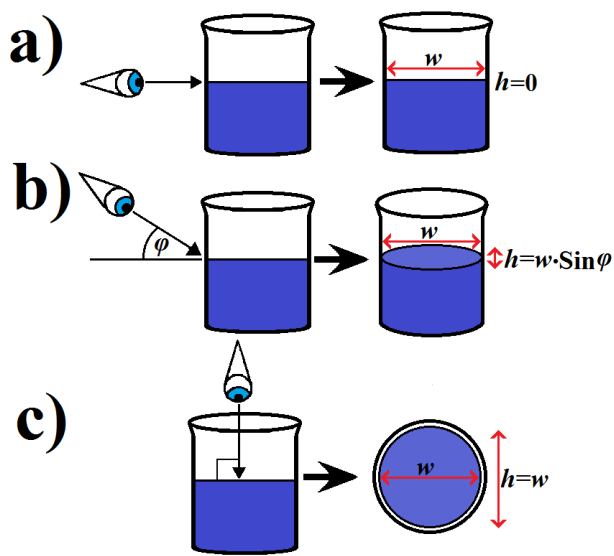

**Figure 5. Effect of angle of view on the observed shape of the surface (for round surfaces). a) Viewed from the same line or from a large distance (angle of view $\varphi=0$), the surface looks like a straight line. b) Viewed from an angle of $\varphi$, the surface looks like a horizontal ellipse of height $h=w\cdot\mathrm{Sin}\varphi$, where $w$ is the ellipse width. c) Viewed from directly above ($\varphi=90°$), the surface looks like a circle.**

Thus, for each horizontal line inside the vessel boundary in the image (Figure 4.b), all of the elliptical curves for which the line is the major axis and the ellipse height ranges from 1 pixel to 30% of the ellipse length are generated (Figure 4.c). The number of scanned curves could be further reduced by ignoring the curves that exceed the container border (Figure 4.c, rightmost). Further reduction could be achieved by ignoring the areas in the vessel that are too narrow relative to the maximum width of the vessel. The last restriction is used because in most vessels, narrow areas correspond to corks and funnels that rarely contain liquids (Figure 1).

**2.2. Rating correspondence between the curves and outlines of liquid surfaces**

Once a curve has been generated, the curve is evaluated and rated by comparing it to the image. The curve is given a score that estimates its overlap with the outline of the real liquid surface in the image. The best method to perform such an evaluation on an elliptical curve is by first dividing the ellipse into upper and lower curves (Figure 4.d). The upper curve contains all of the points in the ellipse that are on or above the center line. The lower curve contains all of the points in the ellipse that are on or below the center line (Figure 4.d). Each of the two curves is matched to the image separately and is given a score that corresponds to the curve overlap with the outline of the liquid surface (Figure 4.e). The best (highest) score of the two curves is used as the full ellipse score (Figure 4.f). This method was found to be much more effective than evaluating the entire ellipse in one step because in the liquid images, half of the liquid surface contour is farther away from the viewer. The shape of this half is therefore more affected by diffraction and absorbance and can have a higher deviation from the elliptical curve, as well as weaker edges. The score of each curve is evaluated by calculating some image property around each point of the curve. This evaluation is discussed in sections 3-4.

**2.3. Selecting the curves threshold score and determining the number of phases**

Once all of the curves have been generated and rated, the next step is to pick the curves that represent real liquid surfaces according to their scores. The major challenge in this step is the determination of a threshold score for accepting a curve as the true outline of the liquid surface in the image. If the vessel is known in advance to contain $N$ liquid phases, it is possible to pick the $N$ curves with the top scores (not including the top and the bottom of the vessel). However, there is often no prior knowledge of the number of separate liquid phases in the image. In such a case, it is necessary to set some threshold score that will determine which curves will be accepted. A good way to pick such a threshold is as some fraction of the best score obtained (for all curves). In this method, the curves are first sorted according to their scores. The threshold score is then taken as a given fraction of the best score (*[Threshold Score]=T·[Best Score]*, 0<*T*<1). This method is effective even if the vessel is empty because, in an image of an empty vessel, the curves corresponding to the top and the bottom of the vessel will still receive relatively high scores that will be used as the threshold.

### 2.3.1. Avoiding multiple recognition of the same surface

The acceptance of more than one curve for a single liquid surface has proven to be a major problem. This problem occurs when the liquid surface in the image convolves with multiple curves and can be partially solved by setting a minimal distance between two accepted curves. If the gap between the two curves is smaller than the minimal distance, the curve with the lower score is deleted.[72]

## 3. Evaluating and rating the correspondence between the curve and the liquid surface line in the image

The ability to evaluate the correspondence between a given curve and a liquid surface in the image is the core of the recognition of liquid surfaces (section 2). This evaluation is performed by examining some property of the image around each point of the curve and using this property to calculate a local score for that point. The curve score could be evaluated by either averaging the local scores for all points on the curve or taking their percentile. The curve score represents a correspondence level between the curve and the outline of the liquid surface in the image. Ideally, the property used to calculate the curve score should be unique to the liquid surface regions in the image. Several image properties have been examined as indicators for the liquid surface and are discussed in section 4. The methods used to calculate the curve score are discussed in section 3.1, below.

### 3.1. Methods for calculating the correspondence score between the curve and the liquid surface in the image

Three computational methods have been used to evaluate the correlation between the curve and the outline of the liquid surface in the image. Detailed step-by-step algorithms of these methods are given in sections 3.1.1-3.1.3 (these sections are not essential for understanding the rest of the paper and can be skipped). Each of the methods could be used to calculate a wide variety of image properties and could be applied to the grayscale image, the gradient map or the edge image (section 1.2), depending on the type of property used as the indicator. The method by which the curve score is calculated has a considerable effect on the result of the recognition process. Similar image properties evaluated using different methods often gave different results in terms of recognition accuracy.

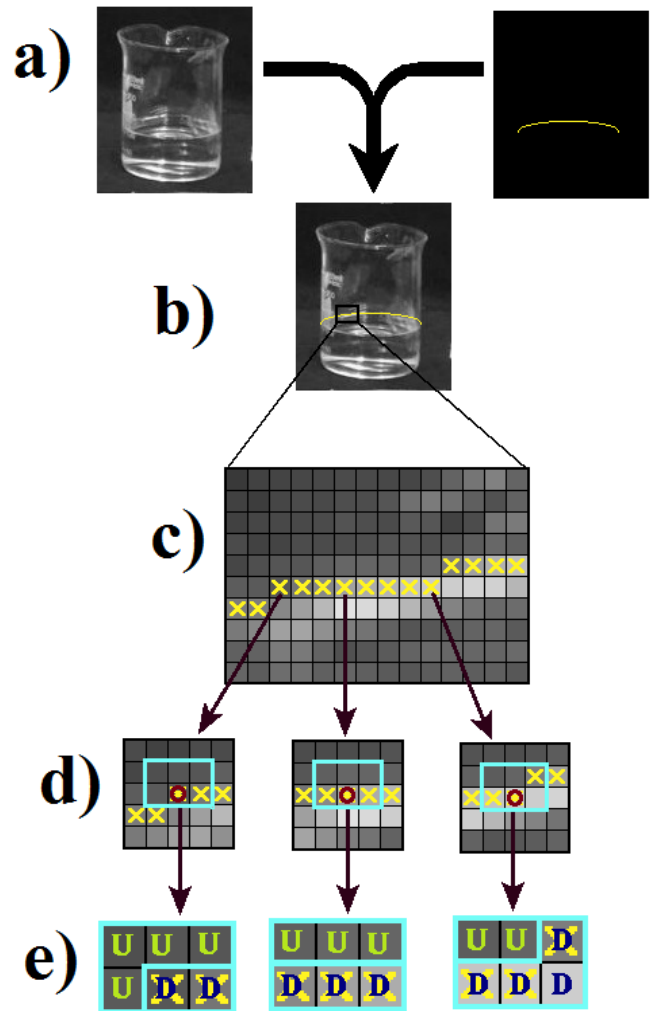

**Figure 6. Method 1: Scoring the match between the curve and the liquid surface using a point-by-point evaluation of the curve surroundings. a-b) For each point in the curve, find the corresponding pixel in the image (the image can be the grayscale, the gradient map or the edge image, depending on the property examined). c) Points in the image corresponding to the curve marked with X. d) Find the region that surrounds each pixel on the curve (the region surrounded by a rectangle).[73] e) The average value of all of the pixels in this region positioned above the curve (marked as U) is calculated as $U$. The average value of all of the pixels in this region positioned on or below this curve (marked as D) is calculated as $D$. The local score of the point is calculated using $U$ and $D$ ((Table 1-2, column 3). The total curve score is calculated as either the average or the percentile of the local scores of all of the points in the curve (Table 1-2, column 6).**

### 3.1.1. Method 1: Point-by-point evaluation of the curve surroundings

Method 1 evaluates the curve correspondence to the liquid surface by evaluating some property of the image in the surroundings of each point of the curve. The method is based on 4 steps (Figure 6):

a) For each point in the curve, find the corresponding pixel in the image (Figure 6.a-c) and the region surrounding this pixel (the region surrounded by a square in Figure 6.d).

b) The average value of all pixels in this region that are located above the curve is calculated as $U$ (U, Figure 6.e). The average value of all of the pixels in this region that are located on the curve or below the curve is calculated as $D$ (D, Figure 6.e).

c) The local score of each curve point is calculated as either $U - D$ (the intensity change normal to the curve) or as $\frac{U-D}{\text{MAX}(U,D)}$ (the relative intensity change normal to the curve). The last equation provided excellent results when used with the grayscale image for the pixel values.

d) Either the average or the percentile of the local scores is used (in absolute value) as the curve score (Section 2.2).

### 3.1.2. Method 2: Direct point-by-point evaluation

Method 2 evaluates the curve correspondence to the liquid surface by examining the values of the pixels that are located directly on the curve. The gradient direction and the normal direction of the curve in this point are used. This approach is implemented in four steps.

a) For each point in the curve, find the corresponding pixel in the image.

b) For each pixel on the curve, evaluate the pixel value as $I$ (Figure 7.c). Additionally, evaluate the curve normal angle $\theta$ (Figure 7.b) and the grayscale image gradient direction $\phi$ at this pixel (Figure 7.a).

c) Calculate the local score as either $I$ or $I \cdot \cos(\theta - \phi)$. The last equation provides good results when using the canny edge image[70] for a pixel value ($I$).

d) Calculate the curve score as the absolute average of the local scores of all of the points on the curve.

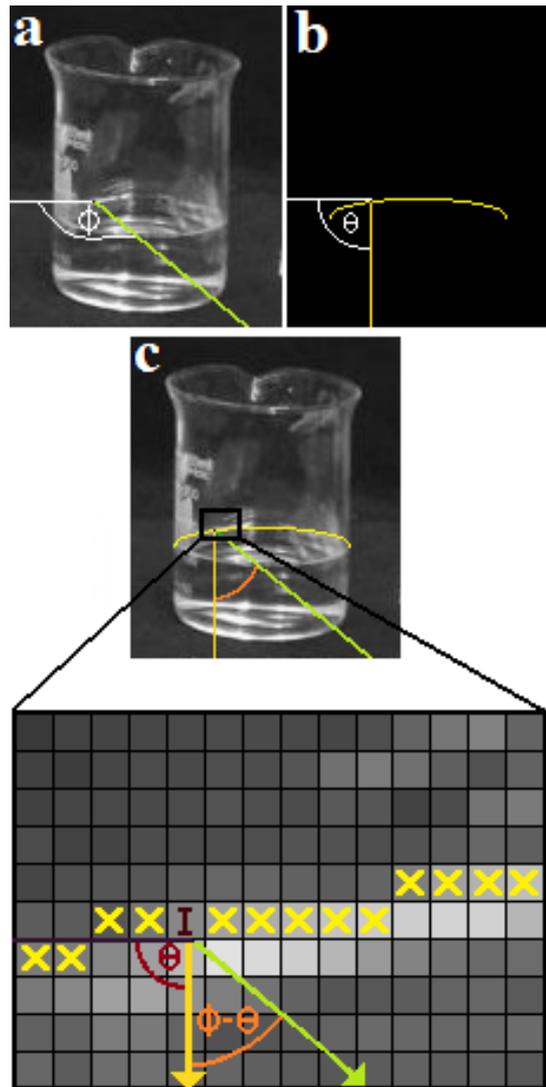

**Figure 7. Method 2 for scoring the match between the curve and the liquid surface using point-by-point evaluation. For each point in the curve, find the corresponding pixel in the image (Panel a-c). Evaluate the pixel value ($I$, panel c), the angle of the curve normal (θ, panel b) and the angle of the grayscale intensity gradient (ϕ, panel a) in this pixel (the curve normal and gradient direction were calculated in a region of 3X3 pixels around each point). Use these three properties to calculate the local score (Table 1-2, column 3). The curve score is the average of the local scores for all of the points on the curve.**

### 3.1.3. Method 3: Regions difference

Method 3 evaluates the correspondence between the curve and the liquid surface in the image by subtracting the average pixel values of one image region from another. The method is based on three steps.

a) Calculate the average value of all of the pixels in the image that are located directly on the curve ($I$ Figure 8.b) as $\bar{I}$.

b) Calculate the average value of all of the pixels that are positioned above the curve and adjacent to the pixels on the curve ($A$, Figure 6.b) as $\bar{A}$.

c) Calculate the curve score as $|\bar{I} - \bar{A}|$ or $\frac{|\bar{I} - \bar{A}|}{\bar{I} + \bar{A}}$. The expression $|\bar{I} - \bar{A}|$ provided good results when used with the canny edge image for pixel values ($A, I$).

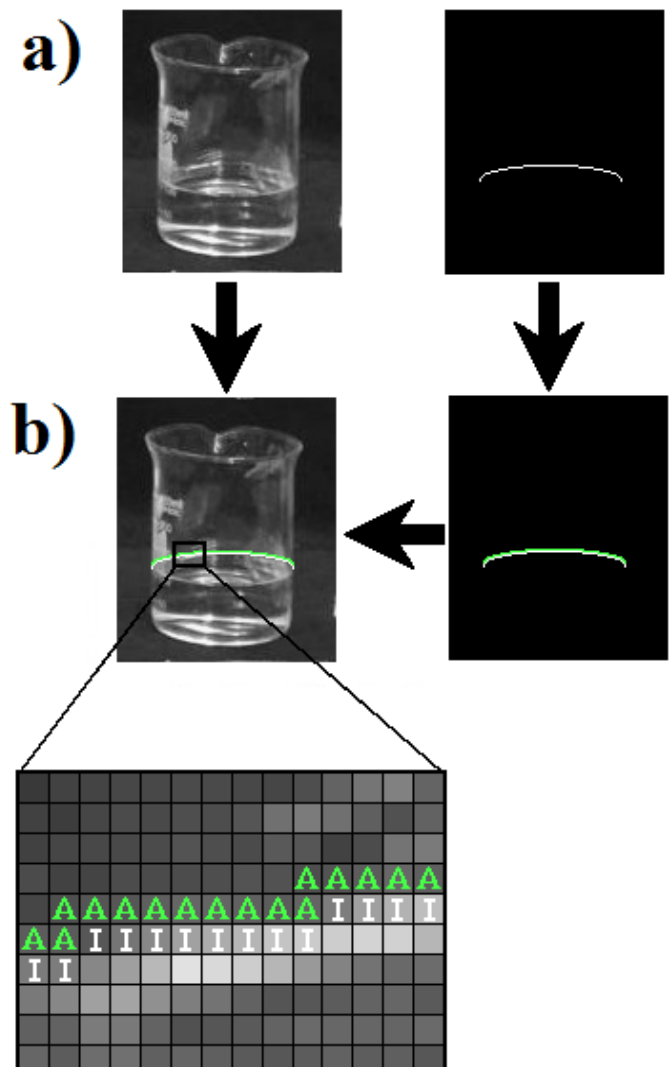

Figure 8. Method 3 for scoring the correspondence between the curve and the liquid surface using the difference between the average pixel values of image regions. a-b) For all points on the curve, find the corresponding pixel in the image (marked $I$) and average the value of these pixels ($\bar{I}$). Find all of the pixels that are adjacent to and above the curve pixels (marked $A$) and calculate their average value ($\bar{A}$). Use $\bar{A}$ and $\bar{I}$ to calculate the total curve score (Table 1-2, column 3).

### 3.2. Experimental procedure

The methods and the indicator for the liquid surface were evaluated based on the test results for 147 images of liquid-containing glass vessels. The source code for all methods is available freely with documentation (See Supporting Information, Section 9). The recognition times were on the scale of 10-60 seconds for most methods (for approximately 200X250 images with an Intel I7 processor). The vessels used for the images were laboratory glassware commonly used in synthetic chemistry, including a beaker, a round bottom flask, an Erlenmeyer flask, a separatory funnel and a chromatography column. Out of 147 total images, 59 contained one liquid, and 88 contained two liquid phases (Figures 1, 9). The liquids used included water, oil, hexane, DMF, THF and silica slurry. The results of each method were evaluated by manually counting the number of the surfaces that were missed and the number of false recognitions of surfaces (Tables 1-2). The pictures of the vessels were taken using a simple digital camera (Canon A810) in the organic chemistry laboratory. Smooth black curtain fabric with no ripples was used as both background and tablecloth.

### 3.2.1. Evaluation of the methods and indicators

Tables 1-2 give the test results for various methods and indicators used to trace the liquid surfaces in the images. The two most important parameters in evaluating the effectiveness of the methods were the miss rate and the false recognition rate (false positives). The miss rate was calculated as the number of surfaces missed divided by the total number of liquid surfaces in all images. The false recognition rate per image was calculated as the number of surfaces that were recognized in the wrong positions divided by the number of images. Both of these parameters and their ratio can be changed by altering the threshold score for accepting the curve (Section 2.3). Therefore, there is no straightforward way to rank various methods for liquid surface recognition. However, it is still possible to identify distinctly better approaches with both a lower miss rate and a lower number of false recognitions.

**Table 1. Results of methods for recognizing liquid surfaces in images, with no consistency check applied** [a]

| Entry | Indicator Description | Local score equation (Score for single curve point) [b,c] | Image Type (Pixel values in local score equation) [d] | Evaluation method of curve score from local scores [e] | Method [f] | Threshold (0,1) [g] | Fraction of missed surfaces for all surfaces (%) [h] | Fraction of missed liquid-air surfaces (%) [h] | Fraction of missed liquid-liquid surfaces (%) [h] | False matches per image (%) [j] | Fraction of correct curve location but wrong shape (%) [i] | Fraction of double recognition of same surface (%) [k] | Miss fraction for surface with emulsion (%) [m] |
|---|---|---|---|---|---|---|---|---|---|---|---|---|---|
| 1 | Relative intensity change normal to curve (percentile) | $\frac{U-D}{\text{MAX}(U,D)}$ | Grayscale | Percentile (65%) | 1 | .4 | 9.8 | 1.4 | 23.9 | 44 | 2.4 | 7.5 | 36.8 |
| 2 | Relative intensity change normal to curve (average) | $\frac{U-D}{\text{MAX}(U,D)}$ | Grayscale | Average | 1 | .4 | 7.2 | 0.7 | 18.2 | 78 | 1.8 | 10.6 | 31.6 |
| 3 | Intensity change normal to curve | $U-D$ | Grayscale | Percentile (65%) | 1 | .3 | 9.4 | 0.7 | 23.9 | 52 | 6.1 | 8.0 | 42.1 |
| 4 | Global relative intensity change normal to curve[74] | $\frac{U-D}{\text{MAX}(\overline{U},\overline{D})}$ | Grayscale | Average | 1 | .4 | 9.4 | 0.7 | 23.9 | 84 | 4.7 | 6.6 | 42.1 |
| 5 | Absolute intensity change normal to curve | $|U-D|$ | Grayscale | Average | 1 | .4 | 11.5 | 1.4 | 28.4 | 87 | 2.9 | 7.2 | 42.1 |
| 6 | Absolute relative intensity change normal to curve | $\frac{|U-D|}{\text{MAX}(U,D)}$ | Grayscale | Average | 1 | .4 | 6.8 | 1.4 | 15.9 | 131 | 5.0 | 7.3 | 21.1 |
| 7 | Average intensity of pixel on curve | $I$ | Grayscale | Average | 2 | .75 | 23.4 | 19.0 | 30.7 | 190 | 4.4 | 0.0 | 47.4 |
| 8 | Relative difference between average intensity above and on curve | $\frac{|\overline{I}-\overline{A}|}{\text{MAX}(\overline{I},\overline{A})}$ | Grayscale | As it is | 3 | .4 | 13.2 | 4.1 | 28.4 | 85 | 8.8 | 5.4 | 47.4 |
| 9 | Normalized difference between average intensity above and on curve | $\frac{|\overline{I}-\overline{A}|}{\overline{I}+\overline{A}}$ | Grayscale | As it is | 3 | .4 | 19.1 | 7.5 | 38.6 | 88 | 11.6 | 10.5 | 57.9 |
| 10 | Difference between average intensity above and on curve | $|\overline{I}-\overline{A}|$ | Grayscale | As it is | 3 | .4 | 14.5 | 4.1 | 31.8 | 63 | 9.0 | 8.5 | 47.4 |
| 11 | Difference between average intensity inside and around surface curve[75] | (Ellipse interior intensity) - (Ellipse outer curve intensity) | Grayscale | As it is | - | .4 | 30.2 | 23.8 | 40.9 | 66 | 1.2 | 1.2 | 42.1 |
| 12 | Relative edge density change normal to curve | $\frac{U-D}{\text{MAX}(U,D)}$ | Edge | Average | 1 | .5 | 10.2 | 3.4 | 21.6 | 59 | 7.1 | 9.0 | 21.1 |
| 13 | Average edge density on curve | $I$ | Edge | Average | 2 | .6 | 11.1 | 5.4 | 20.5 | 63 | 8.6 | 1.0 | 26.3 |
| 14 | Edge density and scalar product of gradient direction and curve normal | $I \bullet \text{COS}(\theta-\phi)$ | Edge | Average | 2 | .45 | 10.2 | 4.8 | 19.3 | 52 | 2.8 | 10.9 | 15.8 |
| 15 | Difference between average edge density above and on curve | $|\overline{I}-\overline{A}|$ | Edge | As it is | 3 | .4 | 7.7 | 2.7 | 15.9 | 72 | 6.5 | 15.2 | 15.8 |
| 16 | Score of the curve in generalized Hough transform (12 angle bins) | Hough transform curve score | Edge | Average | - | .3 | 14.5 | 6.1 | 28.4 | 72 | 7.0 | 1.5 | 26.3 |
| 17 | Change in gradient size normal to curve | $U-D$ | Gradient Size | Average | 1 | .4 | 12.8 | 2.7 | 29.5 | 68 | 3.9 | 6.3 | 52.6 |
| 18 | Average gradient size on curve | $I$ | Gradient Size | Average | 2 | .5 | 15.7 | 6.8 | 30.7 | 79 | 5.1 | 0.0 | 42.1 |
| 19 | Scalar product of gradient and curve normal | $I \bullet \text{COS}(\theta-\phi)$ | Gradient Size | Average | 2 | .5 | 9.4 | 4.1 | 18.2 | 80 | 4.2 | 13.1 | 10.5 |
| 20 | Difference between average gradient size above and on curve | $|\overline{I}-\overline{A}|$ | Gradient Size | As it is | 3 | .4 | 12.3 | 2.7 | 28.4 | 52 | 4.4 | 12.1 | 36.8 |

[a-m] See Table 2 footnotes.

**Table 2. Results of methods for recognition of the liquid surfaces with consistency check applied** [a]

| Entry | Indicator Description | Local score equation (Score for single curve point) [b,c] | Image Type (Pixel values in local score equation) [d] | Evaluation method of curve score from local scores [e] | Method [f] | Threshold (0.) [g] | Fraction of missed surfaces (%) for all surfaces [h] | Fraction of missed liquid-air surfaces (%) [h] | Fraction of missed liquid-liquid surfaces (%) [h] | False matches per image (%) [j] | Fraction of correct curve location but wrong shape (%) [i] | Fraction of double recognition of same surface (%) [k] | Miss fraction for surface with emulsion (%) [m] |
|---|---|---|---|---|---|---|---|---|---|---|---|---|---|
| 21 | Intensity change normal to curve | $U-D$ | Grayscale | Average | 1 | .3 | 9.4 | 0.7 | 23.9 | 67 | 3.3 | 8.5 | 36.8 |
| 22 | Relative intensity change normal to curve | $\dfrac{U-D}{\mathrm{MAX}(U,D)}$ | Grayscale | Average | 1 | .4 | 8.9 | 0.7 | 22.7 | 46 | 2.8 | 6.1 | 42.1 |
| 23 | Global relative intensity change normal to curve [74] | $\dfrac{U-D}{\mathrm{MAX}(\overline{U},\overline{D})}$ | Grayscale | Average | 1 | .4 | 11.1 | 0.7 | 28.4 | 39 | 3.3 | 7.2 | 47.4 |
| 24 | Absolute intensity change normal to curve | $\lvert U-D \rvert$ | Grayscale | Average | 1 | .4 | 11.5 | 1.4 | 28.4 | 44 | 1.4 | 3.8 | 42.1 |
| 25 | Absolute relative intensity change normal to curve | $\dfrac{\lvert U-D \rvert}{\mathrm{MAX}(U,D)}$ | Grayscale | Average | 1 | .4 | 9.4 | 2.0 | 21.6 | 56 | 2.3 | 7.0 | 26.3 |
| 26 | Relative intensity change normal to curve in 1% range (height of curve surroundings is 1% of vessel height) | $\dfrac{U-D}{\mathrm{MAX}(U,D)}$ | Grayscale | Average | 1 | .4 | 10.2 | 0.7 | 26.1 | 57 | 4.3 | 5.7 | 42.1 |
| 27 | Relative intensity change normal to curve in 2% range (height of point surroundings is 2% of vessel height) | $\dfrac{U-D}{\mathrm{MAX}(U,D)}$ | Grayscale | Average | 1 | .4 | 7.2 | 1.4 | 17.0 | 88 | 1.8 | 2.3 | 15.8 |
| 28 | Average relative intensity change normal to curve in the Red, Green, and Blue channels of the RGB color image. | $\dfrac{U-D}{\mathrm{MAX}(U,D)}$ | Color (R, G, B channels) | Average | 1 | .4 | 8.1 | 0.7 | 20.5 | 49 | 5.1 | 7.4 | 42.1 |
| 29 | Edge density change normal to curve | $U-D$ | Edge | Average | 1 | .4 | 10.6 | 2.0 | 25.0 | 61 | 4.8 | 10.5 | 21.1 |
| 30 | Average edge density on curve | $I$ | Edge | Average | 2 | .45 | 8.1 | 2.0 | 18.2 | 57 | 6.9 | 5.6 | 10.5 |
| 31 | Edge density and scalar product gradient direction and curve normal | $I \bullet \cos(\theta-\phi)$ | Edge | Average | 2 | .4 | 8.5 | 3.4 | 17.0 | 46 | 4.7 | 9.3 | 15.8 |
| 32 | Difference between average edge density above and on curve | $\lvert \overline{I}-\overline{A} \rvert$ | Edge | As it is | 3 | .4 | 10.2 | 2.0 | 23.9 | 46 | 5.2 | 9.0 | 21.1 |
| 33 | Scalar product gradient and curve normal | $I \bullet \cos(\theta-\phi)$ | Gradient size | Average | 2 | .4 | 7.2 | 2.7 | 14.8 | 59 | 4.6 | 7.3 | 10.5 |
| 34 | Difference between average gradient size above and on curve | $\lvert \overline{I}-\overline{A} \rvert$ | Gradient Size | As it is | 3 | .32 | 10.6 | 2.0 | 25.0 | 54 | 5.2 | 10.0 | 31.6 |
| 35 | Relative gradient size change normal to curve | $\dfrac{U-D}{\mathrm{MAX}(U,D)}$ | Gradient Size | Average | 1 | .6 | 8.9 | 4.1 | 17.0 | 59 | 2.3 | 2.8 | 15.8 |

[a] Consistency check is used to filter false recognition base on local scores consistency (Section 4.3.1). With consistency check curves were evaluated only if the relative intensity change normal to the curve was larger than 10% for at least 85% of the points (Section 4.3.1).

[b] The equation used to evaluate the local score for each curve point (Sections 3-3.1).

[c] The parameters for the local score equations are discussed in sections 3.1.1-3.1.3 (Figures 6-8). $I$ is the value of the pixel located on the point. $U$ is the average value of the pixels in the point vicinity that are positioned above the curve. $D$ is the average value of the pixels in the point vicinity that are located on or below the curve (Figure 6). $\theta$ is the angle of the normal to the curve at the point. $\phi$ is the angle of the intensity gradient of the grayscale image on the point (Figure 7). $\overline{I}$ is the average value of the pixels located on the curve. $\overline{A}$ is the average value of the pixels adjacent to the curve pixels that are positioned above the curve (Figure 8).

[d] The pixel values used to evaluate the local score (column 3) depend on the image type used for the evaluation (Section 1.2). The pixel values used are intensity (0-255) for the grayscale image, Sobel gradient size (>0) for the gradient image and 0/1 for the edge image (canny, Section 1.2).

[e] The evaluation of the total curve score from the local scores is performed either by averaging the local scores or by taking the percentile of the local scores (Section 4.4). The percentile is either the highest positive value exceeded by 65% of the local scores, or the absolute value of the lowest negative value that is higher than 65% of the local scores (whichever gives the higher result). 'As it is' means that the equation in column 3 is the curve score.

[f] The method gives the algorithm used for the calculation (Section 3.1).

[g] Threshold gives the minimum score that the curve needs to receive to be accepted relative to the best score achieved by the curves in this image ([Threshold Score]=[Max Score]•Threshold, Section 2.3).

[h] The number of liquid surfaces that were missed divided by the total number of liquid surfaces of this type in all of the images (Section 5).

[j] The number of curves that were accepted despite not overlapping with the liquid surfaces in the image, divided by the number of images (Section 6).

[i] The number of cases in which the line of the curve matches the line of a real liquid surface, but the curve shape differs from the shape of the real surface (Section 7).

[k] The number of cases in which multiple curves were identified for a single surface in the image (Section 7).

[m] Miss rate for surfaces with high emulsion (Section 5.2).

# 4. Image properties as indicators for phase boundaries

Several image properties were examined as indicators for the liquid surface. The indicators were evaluated using the methods in section 3.1. The evaluation results appear in Tables 1-2. The image properties that were found to act as the best indicators for the liquid surface are as follows: a) relative intensity change normal to the curve (Section 4.1.3); b) edge density difference between the curve and its surroundings (Section 4.3); and c) combination of the edge density and the scalar product of the curve normal and the gradient direction (Section 4.3.1). The effectiveness of the different indicators is discussed in the following sections.

## 4.1. Intensity, intensity difference and relative intensity difference

Three major properties of the grayscale images (section 1.3) were examined as indicators for the liquid surface boundaries. These properties are the intensity, the intensity change and the relative intensity change (normal to the curve). Of these properties, the relative intensity change was found to be the best indicator for the liquid surface boundaries.

### 4.1.1. Intensity

The edges of the liquid surfaces in the images usually exhibit a high intensity (Figures 9-11), suggesting that the intensity of the pixels along the curve (in the grayscale image) could be used as an indicator for the liquid surface (Entry 7, Table 1). However, the intensity proved to be a poor indicator for the liquid surface because image regions with high intensity also appear in many types of features not related to liquid surfaces (Figures 9-11).

### 4.1.2. Intensity change

The intensity difference is one of the main indicators used by both human and computer vision to identify features in images. The liquid surface in an image usually shows a strong change in intensity normal to the surface boundaries (Figures 9-11), suggesting that the intensity change can act as an effective indicator. The curve local score evaluation was performed using the intensity change normal to the curve around each pixel on the curve (Entry 21, Table 2). This indicator gave good results for surfaces with strong boundaries but weak results for surfaces with weak and blurry boundaries (Entry 21, Table 2).

### 4.1.3. Relative intensity difference

The relative intensity change normal to the curve was found to be the best indicator for the liquid surfaces in the images. This indicator used the intensity change normal to the curve divided by the intensity as the local score (Entries 1, 22, 23, Tables 1-2). The advantage of this indicator is that it balances the illumination effect and the boundary strength, which makes it better for detecting liquid phases with weak boundaries. This approach provided very good results in terms of the recognition and false recognition rates (Entries 1, 22, 23, Tables 1-2). Some of the results of this method are shown in Figure 9.

### 4.2. Intensity change direction and sign consistency

The pixel intensity could either decrease or increase when going from the upper to the lower side of the curve. Therefore, the intensity change normal to the curve could be either positive or negative (Section 4.1). When calculating the curve scores, the sign of the change is not important, and the average change around all of the curve points is taken as an absolute value. However, the consistency of the direction of the change is important. The methods that average the absolute values of the local changes (Entry 5-6 Table 1) provided inferior results compared to the methods that use the absolute value of the average change (Entries 2-3, Table 1). This result can be explained by the fact that intensity changes could result from various image interferences and noises, which tend to have an inconsistent direction. As a result, the intensity changes resulting from noise will tend to cancel each other out when averaged with the sign included.

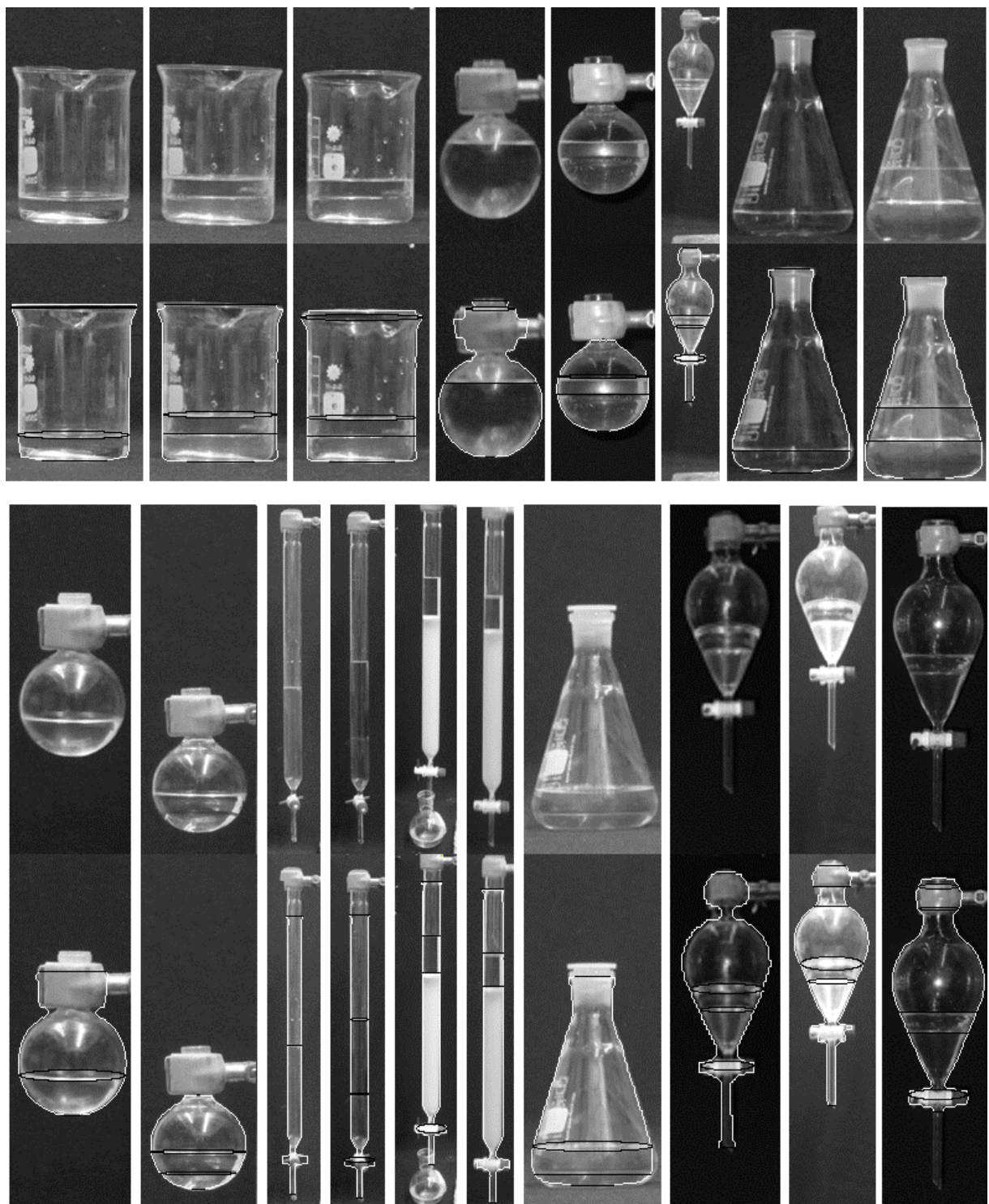

**Figure 9. Examples of successful recognition of liquid surfaces in images, (The results of Entry 22, Table 2). Above: the original image. Below: the image with all recognitions (true and false) marked in black. The floor and ceiling of the vessel outline (marked white) are always marked as the top and bottom phase boundaries.**

### 4.3. Edge-based indicators.

Edges in images are the main property used to identify features by both humans and computers. Edge images are binary images in which the pixels on the feature edges have values of one, while the rest of the pixels have values of zero (Figure 3.d).[68-71] Several edge detectors have been developed for computer vision, the most common of which is the canny algorithm (Section 1.3).[68-71] The liquid surfaces are usually characterized by clear edges in the image (Figure 3.d). We could therefore expect the curves that overlap with the liquid surfaces to have a high correlation with the edges in the edge image. The average edge density on the curve was therefore examined as an indicator for liquid surfaces (average value of the pixel on the curve in the edge image). However, when used alone, this indicator gave only modest results (Entry 30, Table 2). The main problem in this approach is the large number of unrelated edges caused by other features in the image (Figure 3.d). One way to solve this problem is by using the difference between the edge density on the curve and the edge density on the area directly above the curve. This approach prevents false recognition in the noisy image areas with high edge density and results in improved accuracy (Entry 32, Table 2).

### 4.3.1. Combining edge and gradient direction

Another way to improve the accuracy of the edge base indicators is by using the direction of the edge relative to the direction of the curve. The edge line resulting from the boundary of the liquid surface should have the same direction as the curve corresponding to the outline of this surface. To include the edge direction in the calculation of the curve score, the scalar product of the gradient direction (Figure 7) and the curve normal was used (Entry 31, Table 2). This approach provided good results specifically for the recognition of the liquid-liquid surfaces (Entry 31, Table 2). A similar approach is to use the generalized Hough transform for elliptical curves.[68, 69, 76] However, this approach gave inferior results (Entry 16, Table 1).

### 4.4. Average and percentile-based approaches for evaluating curve scores

Methods 1-2 (Section 3.1.1-2) evaluate the correspondence of the curve to the liquid surface by evaluating some image property around each point in the curve. This property is then used to calculate the local score for this point. The curve score could be calculated either by averaging the local scores of all the curve points or by taking the percentile (highest value that is exceeded by 65% of the points). Averaging has the advantage of using the scores of all of the points in the curve. The percentile has the advantage of being more rigid and less sensitive to noise. When used alone, the percentile-based score gave better results (Entries 1, 3, Table 1). However, the best approach was found to be the combination of the percentile and the average of the local scores. This approach uses the average to find the curve score and the percentile as a consistency check for filtering curves with high scores and low consistency. Hence, the score is calculated by averaging the local scores. However, only curves in which the local score of 85% of the points passes the minimal value with a consistent sign are used. This approach was named the consistency check and is discussed in the following section.

**4.4.1. Curve consistency check as the filter for false recognition**

The liquid surfaces in the image exhibit an intensity change that is consistent along most of the surface boundaries (Figures 9-11). Other image features often show a stronger intensity change across their boundaries. However, such features rarely have a shape that is consistent with any possible liquid surface in the image, implying that curves that receive a high score as a result of partial overlap with image features not related to the liquid surfaces could be filtered out by examining their local score consistency. Filtering out curves with high scores but low consistency was found to be highly effective in reducing the false recognition rate. The filtering was performed by using only curves in which 85% or more of the points' local scores have the same sign and values that exceed some minimal threshold (setting a minimal threshold value for the 85$^{th}$ percentile of the local scores). The relative intensity change normal to the curve was found to be a good indicator for such a consistency check (Section 4.1.3). This filter was applied by demanding that at least 85% of the points on a given curve will show a relative intensity change of 10% or more with the same sign.[77]

Adding this filter to various methods considerably reduces the number of false recognitions without any significant increase in the miss rate. This can be seen by comparing the results of the methods that were used with a consistency check (Table 2) and without the consistency check (Table 1). The property used as an indicator for the consistency check does not have to be the same as the image property used to evaluate the curve score. All methods in Table 2 used a consistency check based on the relative intensity change, and all show considerable improvement regardless of the property used to calculate the score (Table 2).

**4.5. Resolution**

The resolution of the scan is essentially the size of the area around each point in the curve where the indicator property was evaluated (Section 3.1.1). For most liquid surfaces, the thickest resolution possible was found to give the best results (examination of the pixels directly on the curve and their direct neighbors, Figures 6-8). However, for dispersive and emulsive phase boundaries (Figure 10), the examination of pixels that are farther away from the curve (2% of the container height) was found to give better results (Entry 27, Table 2, last column) because, in such cases, the phase boundaries are dispersive, and change can occur over a wider range.

### 4.6. Use of color

The curve correlation with the liquid surface was also evaluated using a color image (RGB, Section 1.3). The image was first separated into red, green and blue channels. Each channel was then evaluated by comparing the channel to the curve as if it were a grayscale image with an intensity value that corresponds to the color saturation. The curve scores of the red, green and blue channels were averaged to give the final curve score. However, no real improvement in accuracy resulted from using the color image (Entry 28, Table 2). This could be explained by the fact that the all of the images examined in this work contained transparent liquids or white silica. As result, the color image contained little additional information compared to the grayscale image. For vividly colored liquids, the use of color in the recognition process would likely show superior results.[1]

## 5. Main causes for missed recognition of phase boundaries

Table 2 shows that the recognition of liquid-air surfaces was achieved with high accuracy (less than 10% miss rate) for all evaluation methods. The accuracy for the best methods was nearly perfect, with a miss rate of less than 1%, for liquid-air boundaries (Entries 21-23, Table 2). The boundaries between phase-separating liquids were missed at a much higher rate by all of the methods. This high miss rate is caused by two major properties of liquid-liquid interfaces: a) weak surface boundaries (Figure 10) and b) emulsifying and dispersive phase boundaries. Both are discussed below.

### 5.1. Weak surface boundaries.

Weak phase boundaries are characterized by a small change in intensity across the liquid surface edges (Figure 10). The two main causes of this small change are liquid phases with similar absorbance and refractive indices and high interphase miscibility. Because all of the effective recognition methods utilize the intensity change across the surface boundaries in one way or another, small intensity changes increase the miss rate. However, methods based on the relative intensity change as the indicator are more immune to this effect (Section 4.1.3).

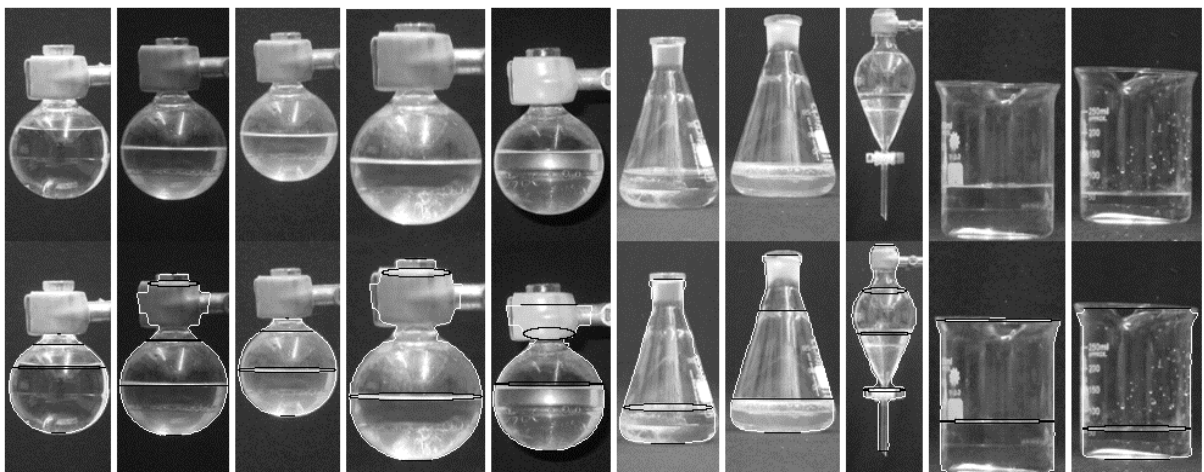

**Figure 10. Examples of missed recognition of the liquid surface as a result of emulsion and weak phase boundaries (based on the method in Entry 22 of Table 2). Above the original image and below the original image with all recognitions and false recognitions marked in black. The floor and the ceiling of the vessel outline (marked white) are always marked as the top and the bottom phase boundaries.**

### 5.2. Dispersive and emulsive surfaces

Emulsion is characterized by droplets of one liquid dispersed in the other liquid. Emulsion occurs as a result of the incomplete phase separation between the liquid phases (Figure 10). Emulsive phase boundaries have two negative effects that disrupt their recognition. First, the emulsion causes the phase transition between liquids to occur over a wide range instead of a thin line. As a result, instead of a sharp intensity change at the phase boundaries, the image shows a weak change over a wide range, making identification more difficult (Figure 10). Another negative effect of emulsion is that the liquid surface is no longer in equilibrium and therefore no longer minimizes its area to form a flat plane. As a result, the liquid surface will no longer take the shape of a line or an ellipse in the image. Therefore, the curves generated during the scan (Section 2.1) can no longer match the liquid surface shape accurately. A partial solution for the recognition of emulsive phase boundaries is using a lower resolution (Section 4.5), i.e., using a larger area in the image above and below the curve line when evaluating the image property used as an indicator for the liquid surface. This practice improves the recognition rate for the emulsive surfaces but reduces the recognition accuracy for all other cases (Entry 27, Table 2, last column).

### 6. False recognition of liquid surfaces

False recognition of a curve as a liquid surface occurs when a curve that does not overlap with any real liquid surface in the image receives a score that passes the acceptance threshold (Figures 11-12). One way to reduce the false recognition rate is by increasing the threshold for accepting the curve as the liquid surface line (Section 2.3). However, increasing the threshold inevitably increases the miss rate of the real surface lines. The major causes of false recognition are image features originating from reflections, labels, sharp turns in the container surface, droplets and nonhomogeneous fluids (Figures 11-12). The edges of these features can overlap with the elliptical curve in the scan (Section 2) and increase its score. As a result, the curve can receive a high score and be accepted even when the curve does not overlap with any real liquid surface in the image. The image features that cause false recognition can be divided into two categories: a) patterns in the image that do not fit the shape of any possible liquid surface outline (Figure 11), and b) patterns with an elliptical or line shape that fits a possible liquid surface (Figure 12). Both categories are discussed below.

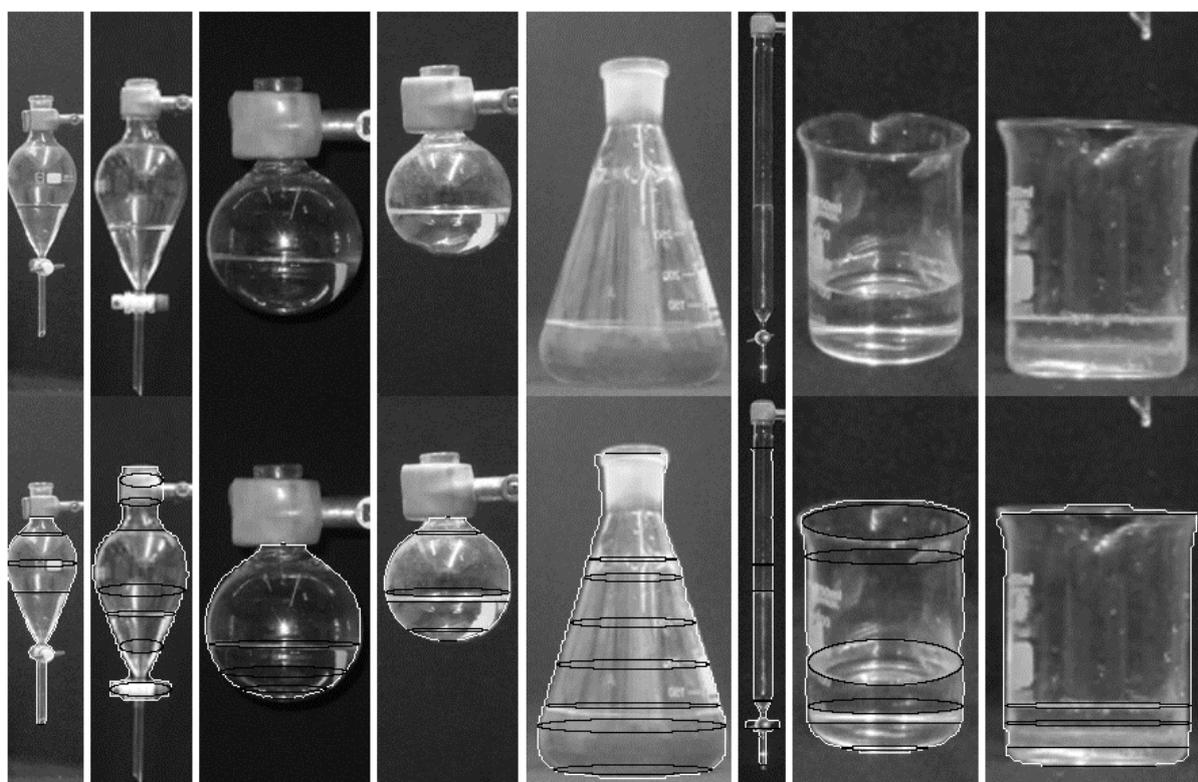

Figure 11. Examples of false recognition of liquid surfaces caused by features in the image that do not resemble the shape of any possible liquid surface. Results of the method in Entry 2, Table 1 (without a consistency check). Above: The original image. Below: The image with all curves that were recognized as liquid surfaces marked in black.

## 6.1. False recognition resulting from patterns that do not fit the shape of any possible liquid surface

Image features that do not fit the elliptical or line shape of possible liquid surfaces can cause false recognition by overlapping with the generated elliptical curve (Section 2.1) in specific areas (Figure 11). The edges of such patterns can be stronger than the edges of the liquid surface boundaries. As a result, even a small overlap between the pattern and a curve can affect the curve score and cause it to be accepted as a liquid surface. However, because such patterns do not follow the shape of any possible liquid surface, their effect might be strong in some areas but inconsistent along the entire curve. Using the consistency of the edges along the curve can therefore be used to filter out false recognitions resulting from such patterns. Percentile-based approaches and consistency check (Section 4.4-4.4.1) have proven to be highly effective for this purpose.

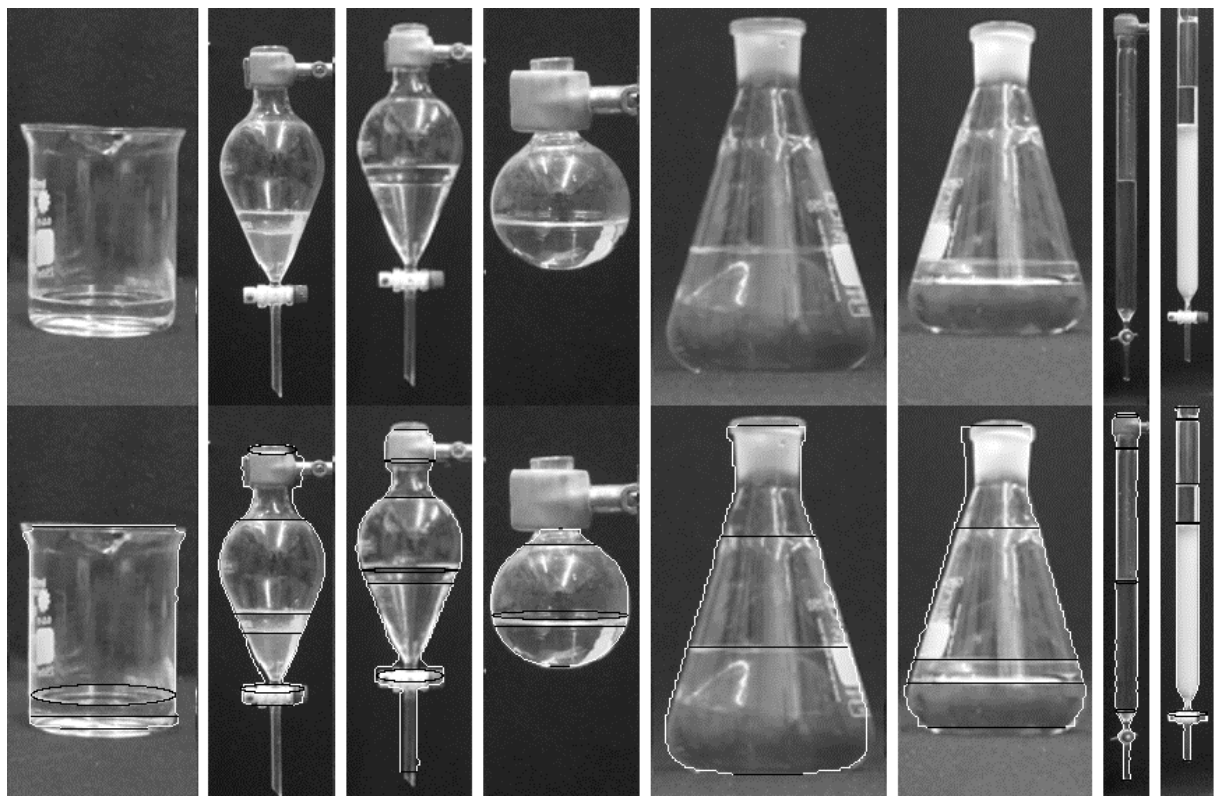

**Figure 12. False recognitions of the liquid surface caused by patterns in the image that follow the outline of a possible liquid surface. Results of the method in Entry 22, Table 2. Above: The original image. Below: The image with all recognitions and false recognitions marked in black.**

## 6.2. False recognitions resulting from patterns in the image that follow the shape of a possible liquid surface

Patterns in an image that have the elliptical or linear shape of a possible liquid surface can mostly be attributed to sharp turns in the vessel contour, glass surface marks and Lambertian reflections[78] (Figure 12). False recognitions resulting from such patterns represent a hard problem, because in terms of shape, they cannot be distinguished from real liquid surfaces. It is possible to determine areas in the vessel that are more susceptible to the appearance of such patterns (for example, parts of the vessel where the surface bends sharply and narrow regions that usually correspond to corks (Figure 12)). However, within the limits of a single image, there is no definitive way to distinguish such patterns from the real liquid surfaces. Such features can likely be handled by using multiple images from different angles or by comparing the image of the empty vessel with that of the filled vessel. These approaches, however, exceed the scope of the current work.

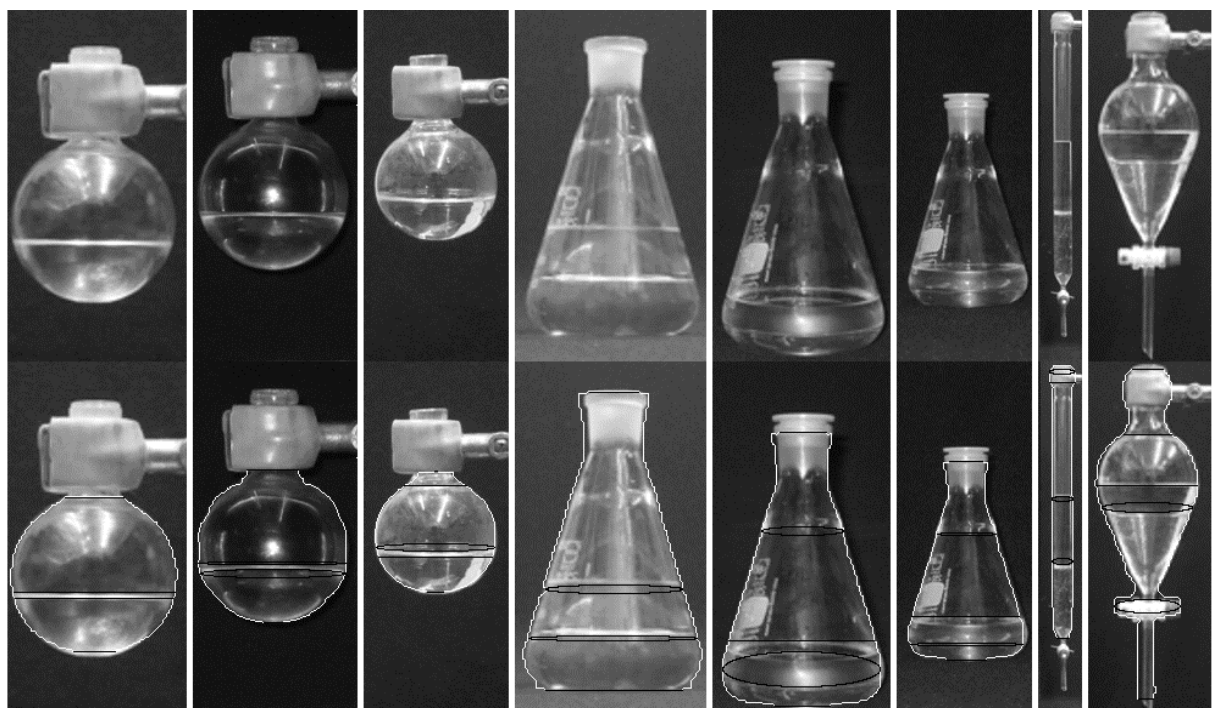

**Figure 13.** Examples for cases with correct recognition of the liquid surface line but with incorrect curve shape or recognition of more than one curve for the same surface. Results of method in Entry 22, Table 2. Above: The original image. Below: The image with all recognitions marked in black.

## 7. Failing to recognize the exact shape of the liquid surface and multiple recognitions of the same surface

Recognition of the liquid surface in the correct location of the image but with the incorrect shape (Figure 13) is a problem that occurs for all of the methods (Tables 1-2). The main cause of this problem is the deviation of the liquid surface shape in the image from a perfect ellipse or line. As a result, no curve generated in the scan (Section 2.1) could accurately match the liquid surface outline in the image. Two causes for distorted liquid surface shapes are capillary force and emulsion (Figure 13, Section 6.2). Another cause is the fact that in the image, one half of the liquid surface outline (the upper or lower half) will be farther away from the viewer (Figure 4.d-e). As a result, its shape in the image will be more distorted by optical interference and will deviate much farther from the shape of an ellipse. Because each half of the curve is evaluated independently (Section 2.3), this phenomenon will not lead to a complete miss of the surface. However, because the recognition is based on the half of the elliptical curve that gives the best match (Section 2.3, Figure 4.e), the other half of the curve will inevitably have a worse match to the surface outline (Figure 13). Another case for which the liquid surface shape deviates from that of an ellipse occurs when the picture of the vessel is taken from a very small distance relative to the vessel radios. This can cause a large deviation between the curvatures the upper half and lower half of the surface ellipse. For such cases it might be necessary to trace the elliptic curve for each half of the liquid surface separately (with different curvature). Multiple recognitions of the same surface is another problem that occurs when more than one curve can be fitted to a given liquid surface. The main cause of this problem is thick surface lines in the image (Figure 13). Requiring a minimal distance between the accepted curves can reduce but not eliminate this problem (Section 2.3.1).[72]

## 8. Summary and Conclusion

This work examined a general computer vision method for the recognition of liquid surfaces in transparent vessels. Such recognition is essential for determining the liquid volume, fill level, phase boundary and the phase separation in various systems. The recognition was performed by first scanning all curves that correspond to possible liquid surface outlines in the image. The number of curves could be limited by assuming the liquid container to be axisymmetric and that the camera and vessel are not tilted left or right. These assumptions limit the possible shapes of the liquid surfaces in the image to horizontal ellipses and lines. Once a curve was generated, it was rated according to its correspondence to a real liquid surface in the image. Dividing the curve into upper and lower halves and evaluating each of the halves separately was found to improve the recognition accuracy. The rating of the curve correspondence to a liquid surface was performed by evaluating some image property around each point of the curve and using the result to determine the curve match score. Several image properties were examined as indicators for liquid surfaces. The best indicators were found to be a) the relative intensity change normal to the curve; b) the edge density on the curve combined with the scalar product of the intensity gradient direction and the curve normal; and c) the difference between the edge density on the curve and the region directly above it. Both the consistency of the indicator along the curve and its average were found to be important in evaluating the curve match to the liquid surface. The indicator consistency (percentile) along the curve was found to be particularly important for filtering out false recognitions. The main causes of false recognitions were found to be image features with the shape of a horizontal line or an ellipse that fit the outline of a possible liquid surface. The recognition of air-liquid surfaces was achieved with very low miss rates (less than 1% miss rate for the good methods). However, the miss rate for the boundaries between phase-separating liquids was much higher (greater than 10% miss rate for all of methods). This higher miss rate can be explained by the emulsions and weak boundaries in such surfaces. The emulsive surfaces were the hardest to recognize due to their unpredictable shape and the blurry boundaries. Using a scan with lower resolution helped to reduce the miss rate for such surfaces. The results of this work suggest that creating a general recognition method for liquid systems is essentially possible and can achieve good accuracy in various cases. However, at this point, no perfect solution has been found with respect to missed recognitions resulting from weak/distorted surfaces or to false recognitions resulting from surface-like features. Additive-based methods for tracing interfaces, such as the colored beads method,[3, 31, 35] remain faster and more accurate due to their ability to circumvent these problems. Further work in this field will require better methods of filtering features resulting from reflections and vessel surfaces, as well as general methods for recognizing phases with non-flat surfaces. From the chemistry perspective, the ability to recognize the phase boundaries of liquids can enable the exploration of various solution phenomena from the machine vision perspective. In addition, such methods can enable the automation of a variety of laboratory techniques.[1, 3, 43-48]

## 9. Supporting Information

a) The full source code for the recognition process described in paper including documentation is available at:

www.mathworks.com/matlabcentral/fileexchange/46893-computer-vision-based-recognition-of-liquid-surface-and-liquid-level-of-liquid-of-transparent-vessel

b) Source code and instruction for the recognition of the boundaries of the transparent vessel (liquid container) in the image are available at:

1) www.mathworks.com/matlabcentral/fileexchange/46887-find-boundary-of-symmetric-object-in-image

2) www.mathworks.com/matlabcentral/fileexchange/46907-find-object-in-image-using-template--variable-image-to-template-size-ratio-

## 10. Referenceand notes

72. Preventing multiple recognition of the same surface is implemented by scanning the accepted curves from higher to lower scores and comparing each curve to all curves with higher scores. If the distance between curves is smaller than some threshold distance, the curve is deleted, and the scan continues to the next curve.
73. The most effective region for local score in method 1(Section 2.2), was found to be the region that contains all the direct neighbors of the pixel, not including the line below the pixel (for the curve corresponding to the upper ellipse half (Region surrounded by rectangle in Figure 6.d)).
74. The global relative intensity change ($\frac{U-D}{\text{MAX}(\bar{U},\bar{D})}$) means that instead of dividing the intensity change by the local intensity maximum ($\text{MAX}(U,D)$), the intensity change is divided by the average intensity maximum along the entire curve ($\text{MAX}(\bar{U},\bar{D})$). $\bar{U}$ is the average value of $U$ for all points on the curve (Figure 6, section 3.1.1). $\overline{D}$ is the average value of $D$ for all points on the curve (Figure 6, section 3.1.1).

75.     The curve score is the difference between the average intensity inside the elliptic curve and the average intensity around the elliptic curve (in a radius of 1 pixel).
76.     Ballard, Dana H. "Generalizing the Hough transform to detect arbitrary shapes."*Pattern recognition* 13.2 (1981): 111-122.
77.     *The consistency check using the relative intensity change is performed by requiring either the relative intensity change of (U-D)/MAX(U, D)>0.1 at 15% percentile or (U-D)/MAX(U, D)<-0.1 at the 85% percentile.*
78.     Pedrotti, F. L.; Pedrotti, L. S.; Pedrotti, L. S., "Introduction to optics." *Prentice-Hall Englewood Cliffs*: 1993; Vol. 2.